\documentclass[lettersize,journal]{IEEEtran}
\usepackage{amsmath,amsfonts}
\usepackage{algorithmic}
\usepackage{algorithm}
\newcommand{\vect}[1]{\mathbf{#1}}

\usepackage{amssymb}
\usepackage{array}
\usepackage[caption=false,font=normalsize,labelfont=sf,textfont=sf]{subfig}
\usepackage{textcomp}
\usepackage{stfloats}
\usepackage{url}
\usepackage{verbatim}
\usepackage{graphicx}
\usepackage{cite}
\usepackage{soul}
\usepackage{color}
\makeatletter
\def\ps@IEEEtitlepagestyle{%
  \def\@oddfoot{\mycopyrightnotice}%
  \def\@oddhead{\hbox{}\@IEEEheaderstyle\leftmark\hfil\thepage}\relax
  \def\@evenhead{\@IEEEheaderstyle\thepage\hfil\leftmark\hbox{}}\relax
  \def\@evenfoot{}%
}
\def\mycopyrightnotice{%
  \begin{minipage}{\textwidth}
  \centering \scriptsize
  Copyright \copyright 20xx IEEE. Personal use of this material is permitted. However, permission to use this material for any other purposes must be obtained from the IEEE by sending an email to pubs-permissions@ieee.org.
  \end{minipage}
}
\makeatother

\begin{document}

\title{SweepEvGS: Event-Based 3D Gaussian Splatting for Macro and Micro Radiance Field Rendering from a Single Sweep}

\author{Jingqian Wu, Shuo Zhu, Chutian Wang, Boxin Shi,~\IEEEmembership{Senior Member, IEEE} and Edmund Y. Lam,~\IEEEmembership{Fellow, IEEE}
\thanks{Jingqian Wu, Shuo Zhu, Chutian Wang and Edmund Y. Lam are with the Department of Electrical and Electronic Engineering, The University of Hong Kong, Pokfulam, Hong Kong SAR, China (e-mail: jingqianwu@connect.hku.hk, zhushuo@hku.hk, ctwang@eee.hku.hk, elam@eee.hku.hk).}

\thanks{Boxin Shi is with the State Key Laboratory for Multimedia Information Processing and National Engineering Research Center of Visual Technology, School of Computer Science, Peking University, Beijing 100871, China (e-mail: shiboxin@pku.edu.cn)}

\thanks{Corresponding author: Edmund Y. Lam.}}



\maketitle

\begin{abstract}
Recent advancements in 3D Gaussian Splatting (3D-GS) have demonstrated the potential of using 3D Gaussian primitives for high-speed, high-fidelity, and cost-efficient novel view synthesis from continuously calibrated input views. However, conventional methods require high-frame-rate dense and high-quality sharp images, which are time-consuming and inefficient to capture, especially in dynamic environments. Event cameras, with their high temporal resolution and ability to capture asynchronous brightness changes, offer a promising alternative for more reliable scene reconstruction without motion blur. In this paper, we propose SweepEvGS, a novel hardware-integrated method that leverages event cameras for robust and accurate novel view synthesis across various imaging settings from a single sweep. SweepEvGS utilizes the initial static frame with dense event streams captured during a single camera sweep to effectively reconstruct detailed scene views. We also introduce different real-world hardware imaging systems for real-world data collection and evaluation for future research. We validate the robustness and efficiency of SweepEvGS through experiments in three different imaging settings: synthetic objects, real-world macro-level, and real-world micro-level view synthesis. Our results demonstrate that SweepEvGS surpasses existing methods in visual rendering quality, rendering speed, and computational efficiency, highlighting its potential for dynamic practical applications.
\end{abstract}




%

\begin{figure}[h]
	
	\centering
	
	\includegraphics[width=0.95\linewidth,scale=1.00]{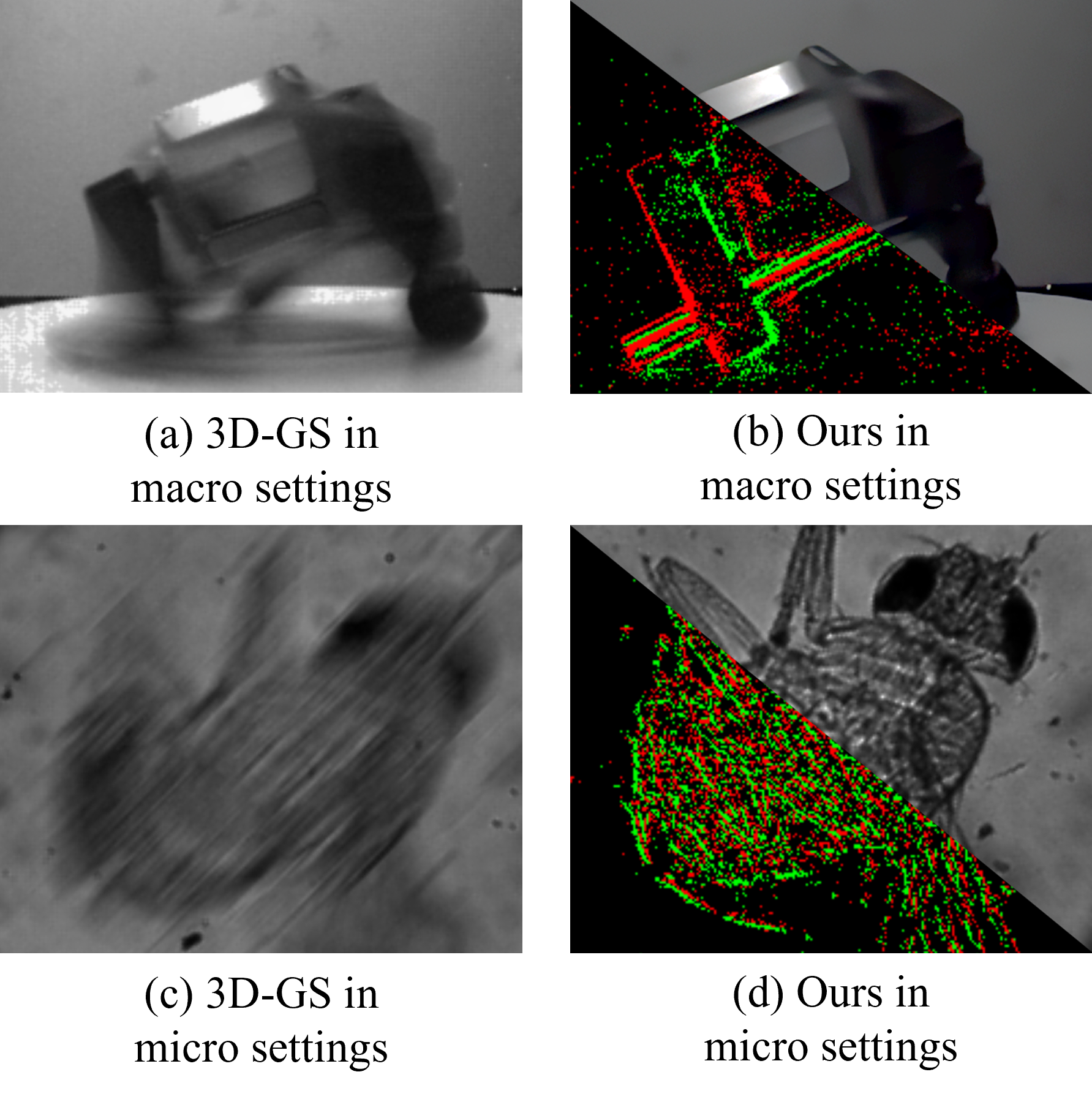}
	
	\caption{Frame-based devices capture experience blur in real-world macro imaging settings when undergoing fast camera/object movement (a). Blur also occurs in real-world microimaging settings such as microscopy when undergoing drastic object displacement (b). Our SweepEvGS is able to render sharp and accurate detailed radiance field representation across various real-world hardware imaging settings (c) and (d).}
	
	\label{fig:vis intro}

\vspace{-0.1cm}

\end{figure}

\section{Introduction}
Novel View Synthesis with densely multi-view images of a given 3D scene generates new and realistic 2D viewpoints \cite{gao2022nerf, zhang2022deep}. Recently, 3D Gaussian Splatting (3D-GS) \cite{kerbl20233d} has introduced an unstructured 3D Gaussian radiance field that uses 3D Gaussian primitives to achieve impressive success in fast, high-quality, and cost-effective novel view synthesis from densely input views \cite{bao20253d, guo2024depth}. However, accurate reconstruction of a radiance field typically requires capturing dense and high-quality training frames. While using videos as input can be an efficient way, blur may occur, producing poor ground truth, shown in Figure \ref{fig:vis intro}(a) and (c), and inaccurate camera calibration \cite{schonberger2016structure, guo2024motion}. Some methods \cite{kerbl20233d, yan2024street, liu2024gva, wen2024gomavatar, deng2021mvf} involve recording videos at a slow pace to minimize motion blur, ensuring that each frame is clear and usable for reconstruction. However, this approach has significant drawbacks. The process is not only slow and cumbersome but also prone to inefficiencies, particularly in dynamic environments where rapid movements are involved. Therefore, efficiently and accurately obtaining such training information has become a vital topic.

Event cameras, on the other hand, are a new imaging sensor that asynchronously captures per-pixel brightness changes \cite{xie2024event}, offering advanced properties and representing a major paradigm shift from conventional frame-based cameras across various vision tasks \cite{zuo2022devo, yang2024latency, zhu2024computational, chen2022ecsnet, liu2022edflow, long2024spike, liu2023motion, zhang2024neuromorphic}. The nature of event cameras makes them highly efficient for fast and dynamic scene capturing. The captured event streams at the microsecond level without motion blur are highly effective for dense radiance field reconstruction. 

These gifts from event cameras make, in theory, good imaging devices for fast imaging and capturing. However, using event cameras to enable robust and accurate real-world radiance field rendering in different imaging settings still faces several technical challenges. 
First, as there are no mature event-based COLMAP \cite{schonberger2016structure} to generate accurate camera pose, coordinate, and point cloud, designing hardware systems that fit different imaging settings while acquiring precise camera calibration for 3D-GS and corresponding clear ground truth frames for quantitative metric calculation is not trivial.
Second, utilizing real-world data and formulating them into accurate supervisory signals across different imaging settings to reconstruct accurate radiance fields is also challenging. This is because object scopes and noise levels differed from each imaging setting, making it harder to unify across all of them.
Third, designing an event-based 3D-GS algorithm for efficient and effective radiance field rendering in these different imaging settings is vital, because such a unified algorithm would benefit brother applications.
To address these challenges, we propose SweepEvGS: a novel hardware-integrated and event-based 3D-GS approach that leverages the capabilities of event cameras for efficient and accurate novel view synthesis across three different imaging settings with just a single sweep of the camera. By using only the first static frame captured and combining it with the dense stream of events generated during the camera sweep, we employ 3D-GS to reconstruct frames of the swept view efficiently and accurately. We summarize our technical contributions as follows:
\begin{itemize}
\item We introduce SweepEvGS, marking the first hardware-integrated approach that enables efficient radiance field rendering from a monocular event camera with only a single camera sweep across different imaging settings.

\item We design a unified and generalized multi-modal data utilization, and robust supervision for 3D-GS pipeline that enables accurate radiance field reconstitution in different imaging settings.

\item We propose different hardware imaging systems designed for various settings of real-world radiance field rendering, including detailed steps for data collection and evaluation for future research. 

\end{itemize}

To demonstrate the robustness of our proposed method, we conducted experiments in three different imaging settings and scales: (1) synthetic objects view synthesis, (2) real-world macro-level view synthesis, for example, Figure \ref{fig:vis intro}(b), and (3) real-world microscopic-level view synthesis, for example, Figure \ref{fig:vis intro}(d). Extensive experiments in both simulated and different real-world settings validate that our method surpasses the visual rendering quality of existing pipeline approaches both quantitatively and qualitatively. Additionally, it achieves around $150$ times faster rendering speeds and requires about $3$ times fewer computational resources than the previous NeRF-based approach. These results demonstrate the robustness and efficiency of our approach in both synthetic and real-world scenarios, highlighting its potential for practical applications in dynamic environments. 

\section{Related Work}

\subsection{Radiance Field Rendering}
Radiance field rendering and novel-view synthesis are fundamental tasks in graphics and computer vision, with significant applications in robotics, virtual reality, and other areas. Neural Radiance Fields (NeRF) \cite{mildenhall2021nerf} and its variants \cite{chen2021mvsnerf, niemeyer2022regnerf} have significantly advanced these tasks by implicitly modeling scenes with MLP-based neural networks and utilizing differentiable volume rendering. This approach achieves state-of-the-art rendering results with high fidelity and fine details. However, the necessity to sample a large number of points to accumulate the color of each pixel results in low rendering efficiency and prolonged training times.

To address these challenges, recent research has explored alternative 3D representations for better efficiency and visual fidelity. 3D-GS employs a set of optimized Gaussian splats to achieve state-of-the-art reconstruction quality and rendering speed. Initialized from sparse Structure-from-Motion (SfM) point clouds, 3D-GS is trained via differentiable rendering to adaptively control the density and refine the shape and appearance parameters. However, accurate radiance field reconstruction usually demands dense, high-quality training frames, which are time-consuming and inefficient to capture. Reducing view constraints can lead to incorrect geometry learning, resulting in poor novel view synthesis and rendering quality. Some methods \cite{kerbl20233d} record videos slowly to minimize motion blur, ensuring clear frames for reconstruction. Yet, this approach is slow, cumbersome, and inefficient, especially in dynamic environments with rapid movements. Therefore, it is vital to seek solutions that overcome these challenges and synthesize novel views with simple imaging processes across different settings.

\subsection{Neuromorphic Radiance Field Rendering}
The distinctive features of event cameras, including their ability to prevent motion blur, provide a high dynamic range, ensure low latency, and consume minimal power, have spurred their adoption in computer vision and computational imaging \cite{han2023hybrid, duan2023neurozoom, zhang2024joint, zhu2024efficient, wang2024neuromorphic}. Several approaches have been proposed to address the view synthesis challenge using NeRF \cite{mildenhall2021nerf} with event data \cite{klenk2023nerf, rudnev2023eventnerf}, leveraging volumetric rendering with either pure event or semi-event (blurred RGB involved) supervision. These methods utilize the inherent multi-view consistency of NeRFs, providing a robust self-supervision signal for extracting coherent scene structures from raw event data. However, a significant drawback is the time-consuming optimization of NeRF. The computational demands of training and optimizing an event NeRF pipeline, in terms of both training time and GPU memory, are approximately 80 times greater than those of the 3D-GS pipeline. Additionally, the use of high-dimensional multilayer perceptron networks in the NeRF architecture results in a view-rendering speed that is around 190 times slower, which limits its applicability for real-time rendering.

\begin{figure*}[t]
\begin{center}
  \includegraphics[width=1 \linewidth]{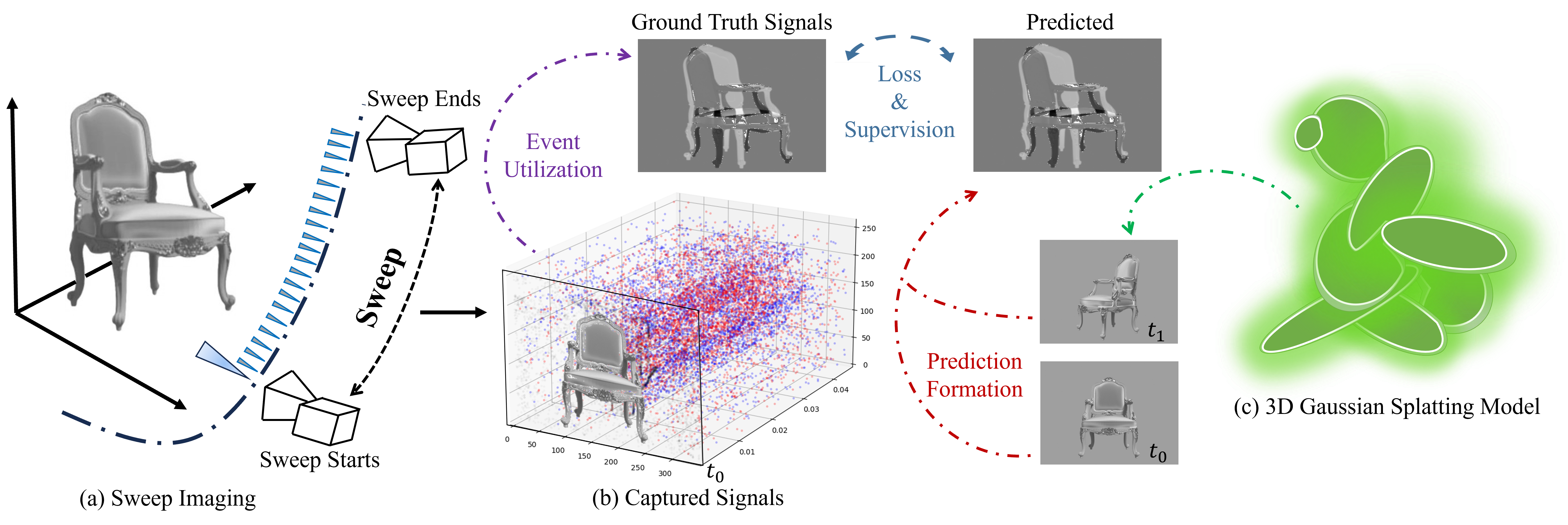}
\end{center}
   \caption{Overview of the SweepEvGS pipeline for novel view synthesis. The process begins with capturing a static object or scene using an event camera (a). The camera captures signals (b) including the initial frame before the sweep starts when static, and the dense event streams representing intensity changes during the camera sweep. On the one hand, these signals are utilized to form (purple arrow) the ground truth signals for training and supervision, as described in Sec. \ref{Sec: Event Stream Utilization}. On the other hand, the 3D-GS model, described in Sec. \ref{Sec: 3D GS}, renders (green arrow) when given the corresponding camera position and pose, and formalizes (red arrow) the rendered results to predictions. The designed loss supervises the training pipeline, as described in Sec. \ref{Sec: Supervision}. Our method efficiently bridges the gap between sparse event data and dense scene reconstruction, enabling high-quality novel view synthesis.}
\label{fig:pipline}
\end{figure*}

A few recent works have attempted to overcome these efficiency issues by combining 3D-GS with event data. \cite{wang2024evggs} proposed a collaborative learning framework for event depth estimation, event-to-frame reconstruction, and event-based 3D-GS. However, this approach requires prior features and outputs from previous depth estimation and frame reconstruction tasks. The entire training framework also necessitates ground truth depth maps and frame references, limiting its usage, especially in scenarios where only event streams are available. \cite{wu2024EV-GS, han2024event} proposed a pure event-based 3D-GS end-to-end method for novel view synthesis. Nonetheless, the rendering quality was not satisfactory as the work was mainly designed for synthetic event data, and the method did not explore different real-world hardware-involved imaging setups.

\subsection{Radiance Field Rendering with Real-World Data}
Motivated by the potential of radiance field rendering and novel view synthesis, extensive work has been conducted on real-world datasets. NeRF has been widely applied to real-world object-level and scene-level rendering \cite{mildenhall2019local, mildenhall2021nerf, barron2022mip}. Additionally, NeRF has been utilized in microscopy imaging settings \cite{zhou2023fourier}. Although 3D-GS has not yet been applied to microscopy and other optical settings, it has been successfully used for real-world object and scene rendering \cite{kerbl20233d, yu2024mip, gao2023relightable}. Currently, there has been no research exploring the application of real-world event data with 3D-GS across various real-world imaging systems. To address this gap, our work examines the use of 3D-GS with event data in three different levels of imaging setups. We conduct both quantitative and qualitative analyses, and we provide detailed information about our different hardware setups for real-world data collection. we hope to provide valuable insights for future research in this area.

\section{Methodology}
\subsection{Preliminary on 3D Gaussian Splatting}
\label{Sec: 3D GS}
3D-GS \cite{kerbl20233d} represents detailed 3D scenes using point clouds, where Gaussians are used to define the structure of the scene. Each Gaussian is characterized by a 3D covariance matrix $\Sigma$, and a central point $\vect{x}$: 

\begin{equation}
    G(x) = \exp \left(-\frac{1}{2}\vect{x}^{T}\Sigma^{-1}\vect{x} \right),
    \label{GS_eq1}
\end{equation}
where the central point $\vect{x}$ is referred to as the mean value of the Gaussian. To facilitate differentiable optimization, the covariance matrix $\Sigma$ can be decomposed into a rotation matrix $R$ and a scaling matrix $S$:

\begin{equation}
    \Sigma = RSS^TR^T.
\end{equation}

To generate renderings from different viewpoints, the splatting technique described in \cite{yifan2019differentiable} is used to project the Gaussians onto the camera planes. This method, initially introduced in \cite{zwicker2001surface}, involves a viewing transformation matrix $W$ and the Jacobian $J$ of the affine approximation of the projective transformation. Using these, the covariance matrix $\Sigma'$ in camera coordinates is given by:

\begin{equation}
    \Sigma' = J W ~\Sigma ~W ^{T}J^{T}.
    \label{Eq: GS}
\end{equation}

To summarize, each Gaussian point in the model is defined by several attributes, i.e., its position $\vect{x} \in \mathbb{R}^3$, color represented by spherical harmonics coefficients $\vect{c} \in \mathbb{R}^k$ (where $k$ denotes the degrees of freedom), opacity $\alpha \in \mathbb{R}$, a rotation quaternion $\vect{q} \in \mathbb{R}^4$, and a scaling factor $\vect{s} \in \mathbb{R}^3$. For each pixel, the color and opacity of all Gaussians are computed based on the Gaussian representation outlined in Equation \ref{GS_eq1}. The blending process for the color $C$ of $N$-ordered points overlapping a pixel follows this formula:

\begin{equation}
    C = \sum_{i \in N} \vect{c}_i \alpha_i \prod_{j=1}^{i-1} (1-\alpha_j),
\end{equation}
where $\vect{c}_i$ and $\alpha_i$ denote the color and density of a specific point, respectively. These values are influenced by a Gaussian with a covariance matrix $\Sigma$, which is subsequently adjusted by per-point opacity and spherical harmonics color coefficients.

\subsection{Noisy Events as 3D-GS Supervisory Signals}
\label{Sec: Event Stream Utilization}
Each event $e_k$ is described as a tuple $(\vect{x}_k, t_k, p_k)$, which occurs asynchronously at pixel $\vect{x}_k = (x_k, y_k)$ at a microsecond timestamp $t_k$. The polarity $p_k \in \{ -1, +1 \}$ denotes either an increase or decrease in the logarithmic brightness $L(\vect{x}_k, t_k)$ by the contrast threshold $A$. Specifically, an event at time $t_k$ is triggered if the following condition is met:

\begin{equation}
    \Delta L \triangleq L(\vect{x}_k, t_k) - L(\vect{x}_k, t_{k-1}) = p_kA,
\end{equation}
where $t_{k-1}$ is the timestamp of the previous event at pixel $\vect{x}_k$. As we utilize event stream in a unifying way across all imaging settings and microscopic settings tend to introduce more noise \cite{yang2023sci}, we apply the $y$-noise filter from \cite{feng2020event} for preprocessing noisy event. Noise filtering is vital for accurate view reconstruction \cite{duan2021guided}, as illustrated in the supplementary material. As illustrated in Figure \ref{fig:pipline}, our goal is to render a radiance field representation using differentiable 3D Gaussian functions under pure event signal supervision with only the starting frame for camera calibration, without involving other frame-based data. To achieve this, we need to convert the ground truth event data into a differentiable supervisory signal and train a 3D-GS model to render this representation. Our algorithm generates a rendering result at $t_k$ with camera pose $p_k$ the timestamp where the camera is currently at. With the known initial frame at timestamp $t_0$ with camera pose $p_0$, meaning before the camera begins to sweep, the supervisory signal is formed by accumulating ground truth event signal between those two timestamps.

For each timestamp $t_k$, we calculate two associated rendering results from the 3D Gaussian model: $I_k = G(p_k)$ and $I_0 = G(c_k)$, where $I_k$ and $I_{0}$ are the rendered grayscale frames at times $t_0$ and $t_k$, respectively, $G$ is the 3D-GS model, and $p_k$ and $p_0$ are the camera poses at times $t_0$ and $t_k$. We represent the logarithmic image as $L(I_t) = \log\big((I_t)^g + \epsilon\big)$, where $\epsilon = 1 \times 10^{-5}$ (to avoid NaN values), and $g$ denotes a fixed gamma correction value set to $2.2$ across all experiments. This conforms to the grayscale gamma curve, with Gamma $2.2$ serving as the recommended smooth approximation \cite{international1999multimedia, rudnev2023eventnerf}. Consequently, we derive the predicted accumulative difference $E_{pred} = L(I_0) - L(I_k)$.

To utilize event data as a supervisory signal for $E_{pred}$, following \cite{zihao2018unsupervised}, we aggregate the polarities of all events occurring between the selected times $t$ and $t-w$ based on their positional information $\vect{x}_k$, such that

\begin{equation}
    E_{gt} = \int_{t}^{t+w} e_{k}(\vect{x}_{k}, p_{k}, t_{k}) \, dt,
    \label{Eq: gt}
\end{equation}
where $E_{gt}$ is the aggregated result.

\subsection{Supervising Events to Radiance Fields}
\label{Sec: Supervision}
Pure events as input give poor real-world radiance field rendering results \cite{wu2024EV-GS}. Since events represent the intensity changes from previous timestamps, the clear and shaped initial frame captured by the static camera before the sweep is valuable knowledge of where the events changed from. Given a camera pose at a specific time $t$, the 3D-GS model learns and renders the corresponding frame $I_t$ with supervision. Because the first frame $I_0$ is known for camera calibration, our task is to formulate the accumulated difference $E_{pred}$ between $I_t$, $I_0$, and supervise it with the ground truth event $E_{gt}$. To effectively supervise $E_{pred}$ from $E_{gt}$, we follow the methods described in \cite{delbruck2020v2e, wu2024EV-GS, klenk2023nerf}, applying the linlog mapping to derive the predicted logarithmic brightness difference for both $E_{pred}$ and $E_{gt}$. Using a normalized $\mathcal{L}_2$ loss, the loss $\mathcal{L}_{event}$ is calculated as:

\begin{equation}
    \mathcal{L}_{event}(x, y) = \frac{\sum\limits_{i=1}^H \sum\limits_{j=1}^W \big(\| L(x) \|^2_2 - \| L(y) \|_2^2\big)^2}{H \times W},
\end{equation}
where for an arbitrary $u$:

\begin{equation}
    L(u) = \mathop{\mathrm{linlog}} \big(I(u)\big) \triangleq
    \begin{cases}
        I(u) \times \ln(B) / B & \text{if } I(u) < B, \\
        \ln\big(I(u)\big) & \text{otherwise},
    \end{cases}
\end{equation}
where the threshold $B$ delineates the linear region where no logarithmic mapping is applied. The value of $20$ is used for $B$ in all our experiments. Here, $x$ and $y$ correspond to $E_{pred}$ and $E_{gt}$, respectively.

Additionally, we retained the D-SSIM loss used in the original 3D-GS approach, as we found that a small coefficient improves the rendering result. Following previous works, we calculate the D-SSIM loss as:

\begin{equation}
    \mathcal{L}_{D-SSIM}(x, y) = \frac{(2\mu_{x}\mu_{y} + c_1)(2\sigma_{xy} + c_2)}{(\mu_{x}^2 + \mu_{y}^2 + c_1)(\sigma_{x}^2 + \sigma_{y}^2 + c_2)},
\end{equation}
where $\mu$ is the pixel sample mean, $\sigma_x$ is the variance of $x$, $\sigma_{xy}$ is the covariance of $x$ and $y$, $c_1 = (k_1q)^2$, and $c_2 = (k_2q)^2$ are constants to stabilize the division with a weak denominator. $q$ is the window range of the pixel values, and $k_1$ and $k_2$ are constants. 
Therefore, the loss function is derived as follows:

\begin{equation}
\begin{split}
    \mathcal{L} = \mathcal{L}_{event}(E_{pred}, E_{gt}) + \\ \lambda (1 - \mathcal{L}_{D-SSIM}(E_{pred}, E_{gt})),
\end{split}
\label{Eq: loss}
\end{equation}
where $\lambda$ is set to $0.1$ for all our experiments.

Our supervision method is designed to be robust across various types of data, including synthetic, real-world macro, and real-world microimaging data. Introducing the $y$-noise filter eliminates interference signals, especially for microimaging data collected under the microscope. Incorporating normalization techniques ensures the loss remains stable across different imaging conditions. This stability is crucial for maintaining consistent performance and achieving high-quality rendering results, regardless of the data source. More ablation studies in the supplementary materials support our claim.

\section{Experiments}
\subsection{Overview}
\label{Exp Overview}
This subsection provides an overview and structural layout of our experiments. Here, we briefly introduce the experiment setups. Experiments are conducted in three different imaging settings to demonstrate the robustness of our approach: 1) real-world macro-level sequences, 2) real-world microscopic-level sequences, and 3) synthetic sequences.

\begin{figure}[t]
	\centering
	\includegraphics[width=\linewidth, scale=1.0]{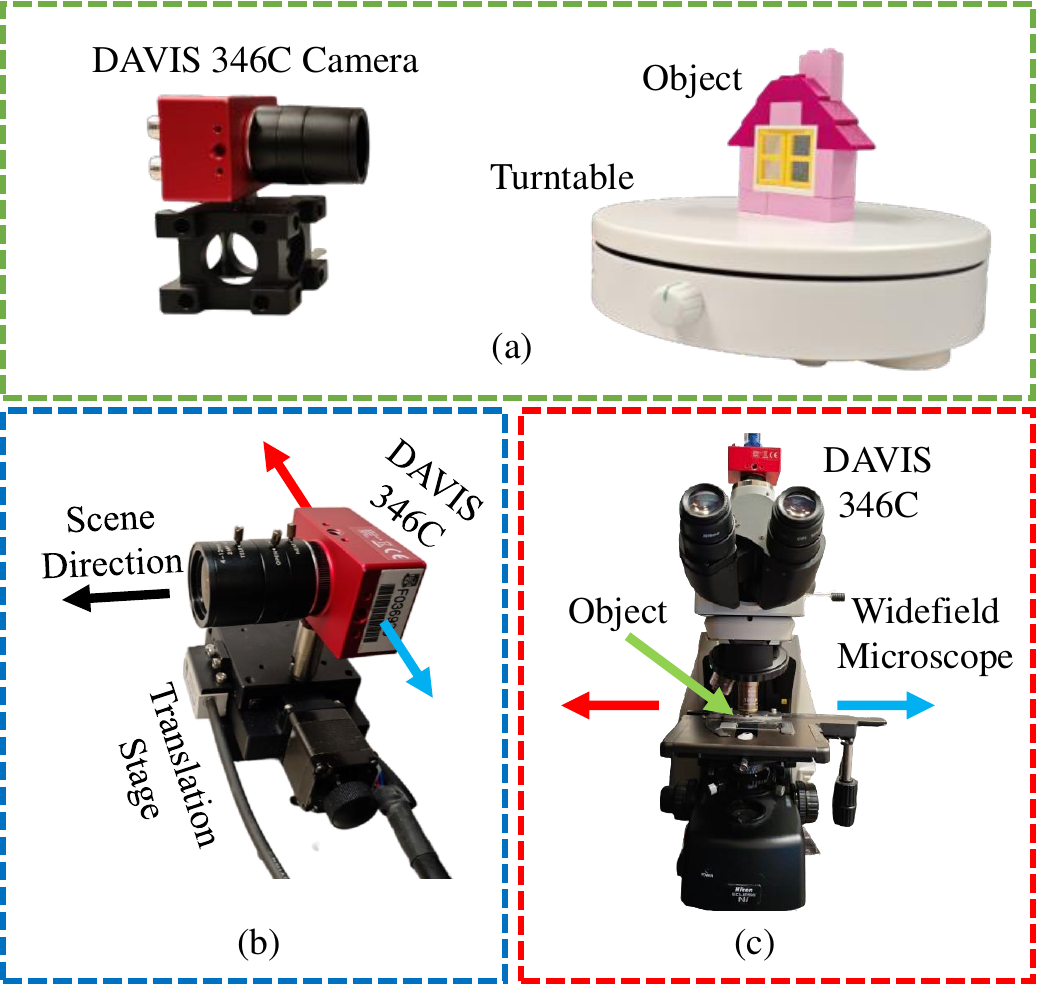}
	\caption{Demonstration of Hardware Setup. (a) Real-world macro camera system setup for objects with depth:  the DAVIS 346C event camera is fixed, and the object is placed on a fast turntable with known constant speed. The camera will capture events when there are changes in the view angle from the object. (b) Real-world macro camera system setup for plain objects: the DAVIS 346C event camera is placed on a translation station that enables horizontal linear movements in both red and blue directions. (c) Real-world microscopic camera system setup: the DAVIS 346C event camera is placed on a widefield microscope that enables horizontal linear movements in both red and blue directions. The camera will capture events and frames from the object placed on the observation plate.}
	\label{fig:setup}
\end{figure}

All real-world data were captured using the DAVIS 346C event camera. For real-world macro sequences, we introduce two types of imaging systems for different scenarios. For objects that involve depth information (such as a Lego toy and an optical component), we fixed the event camera, and ``swept'' the objects using a fast turntable with a known constant speed, shown in Figure \ref{fig:setup}(a), to generate accurate camera poses and coordinates.
For objects that are plain (such as a keyboard and a cloth), we mounted the event camera on a horizontal translation stage, shown in Figure \ref{fig:setup}(b), allowing linear movement at a constant speed to generate accurate camera poses and coordinates. For quantitative comparison, we captured two sets of data for each sequence: (1) a fast sweep (high-speed mode) to capture the event stream and RGB frames generated by the DAVIS 346C camera, and (2) a slow sweep (low-speed mode) to capture clear and sharp RGB frames used as ground truth references, since the fast sweep results in blurry RGB frames unsuitable for quantitative comparison. We aligned the time correspondence between these datasets to accurately calculate quantitative metrics.

For the real-world microscopic-level imaging setup, we deployed the DAVIS 346C camera on a microscope station, using the station's mobility control to ``sweep'' the object horizontally or vertically. Due to manual control of the object's movement, we collect the event stream and RGB frames under a fast sweep and clear and sharp RGB frames under a slow sweep with a static frame capturing process to collect the RGB ground truth references. We then manually paired the fast-slow data as in the other setups. Thus, we could also compare the visual rendering quality between the rendered result and a reference image of the object taken statically under quantitative metrics. Figure \ref{fig:setup} shows a photograph of the setup, and detailed hardware information will be presented in Sec. \ref{Sec: Headware}.

For the synthetic sequences, we used 3D models from \cite{mildenhall2021nerf} to simulate the sweep process of a camera in a real-world setting. Each scene was captured by sweeping the camera 90 degrees around the object, resulting in 250 RGB images, which were then converted to grayscale. An event stream was generated from these images using the model from \cite{gehrig2020video}, and we applied the corresponding camera intrinsic and extrinsic in our approach.

The rendering results of different approaches are evaluated using the peak signal-to-noise ratio (PSNR) and the structural similarity index measure (SSIM). PSNR assesses signal fidelity by comparing the maximum signal power to noise power, while SSIM evaluates image similarity based on luminance, contrast, and structure. These metrics provide quantitative insights into the quality and fidelity of the rendered images.

While the above provides an overview of the experiment setup, detailed hardware information and implementation details of the algorithm will be presented in Sec. \ref{Sec: Headware} and Sec. \ref{Sec: Implementation}. The results of the three types of experiments mentioned will be presented in Sec. \ref{sec:real_world_experiment}, Sec. \ref{Sec: real-world micro}, and Sec. \ref{Sec: Synthetic}. We also conduct an ablation study in Subsection \ref{Sec: Ablation} to demonstrate the necessity and effectiveness of our method design. Last, qualitative analysis of the methods and results will be presented in Sec. \ref{Sec: analysis}.

\begin{figure*}[t]
\begin{center}
  \includegraphics[width=1 \linewidth]{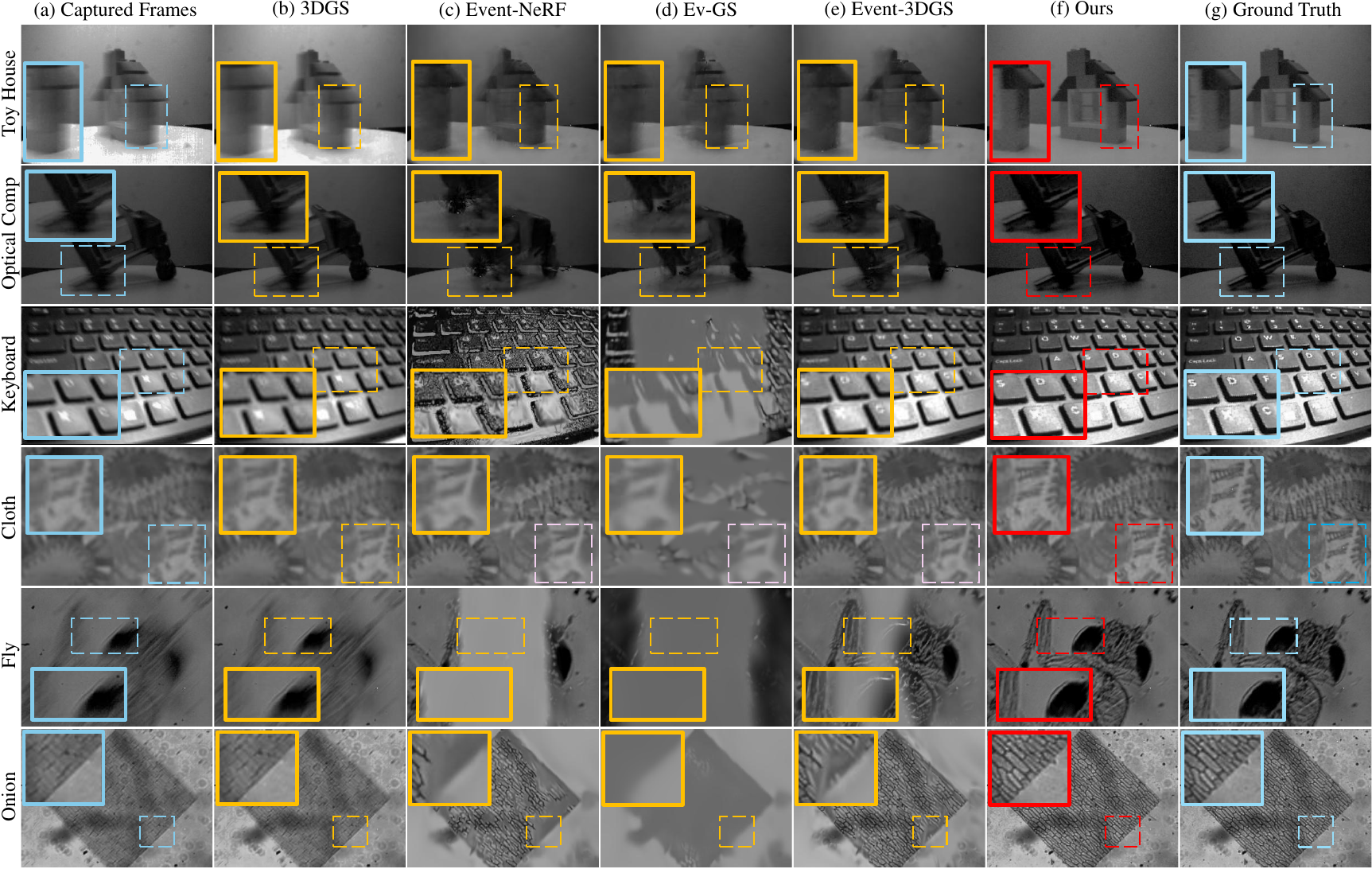}
\end{center}
   \caption{A visual comparison of our approach against others on real-world datasets: the captured frames (a) from traditional frame-based camera results in motion blur; using traditional frame-based radiance field approach, such as 3D-GS (b) produce blurred radiance field reconstruction and view rendering; event-based NeRF approach (c), time and resource consuming while still produce accurate rendering results; further, existing event-based GS approaches (d and e) failed on real captured data as more noise and randomness are introduced in event streams; our approach (f), leveraging the proposed multi-modal data utilization and supervision, renders sharp and accurate results compared to the ground truth references (g). We visually compare the results across all methods on real-world macro sequences (toy, optical component, keyboard, and cloth) captured by hardware in Figure \ref{fig:setup}(a) and Figure \ref{fig:setup}(b); and also on real-world microscopy sequences captured by hardware in Figure \ref{fig:setup}(c). Dotted boxes indicate areas of interest, and the solid line boxes reveal zoomed results.}
\label{fig:vis all}
\end{figure*}

\subsection{Hardware Detailed Information}
\label{Sec: Headware}
This subsection presents the detailed hardware information used in the real-world data collection phase to contribute to future research. A video is included as supplementary material to demonstrate the hardware setups used in this paper.

\begin{itemize}
    \item \textbf{Event Camera}: Shown in Figure \ref{fig:setup} (a), in all real-world experiments, we use DAVIS 346 Colorful event camera, a state-of-the-art imaging device designed to capture event stream and frame data. With a frame resolution of 346 pixels by 260 pixels, this camera offers a detailed view of the environment. One of the key features of the DAVIS 346 C is that it can react to events in the scene with microsecond precision. This makes it particularly effective for capturing fast-moving objects and dynamic events that would typically be blurred or missed by conventional cameras.

    \item \textbf{Translation Stage}: Shown in Figure \ref{fig:setup} (b), in real-world macro experiments, we use WNMC400 motion controller to provide precise camera trajectory control. The WNMC400 motion controller is a compact and high-performance device designed for precise control of industrial automation systems. It offers robust motion control capabilities with features such as trajectory planning, electronic gearing, and synchronization, making it ideal for applications requiring accurate and coordinated motion. With its unique and precise control, we set the speed of the fast motion (input only) to 8 times faster than the speed of the slow motion (ground truth only). The known trajectory and speed relationship makes it easy for us to generate precise global coordinates and frame correspondence needed for metric computation.

    \item \textbf{Microscope}: Shown in Figure \ref{fig:setup} (c), in real-world microscopic experiments, we use Nikon Model ECLIPSE Ni-U Microscopy equipment to observe microscopic data. The Nikon Model ECLIPSE Ni-U is an advanced research-grade microscope known for its exceptional optical performance and ergonomic design. In the context of microscopy, minor shifts in the specimen or the camera can lead to considerable displacement. Such movements can render frame-based cameras ineffective, producing blurred images. Consequently, this constraint impedes conventional radiance field rendering techniques from accurately reconstructing clear and detailed textural information within microscopic environments. We manually control the movement of objects using the microscope with linear movement, making it easy to generate global coordinates. However, we did not find a way to provide precise control of the speed like the macro setup, therefore, we manually aligned the fast sweep frame and the slow sweep ground truth frame to compute quantitative metrics.

\end{itemize}

\subsection{Implementation Details}
\label{Sec: Implementation}
Our implementation is based on the 3D-GS codebase \cite{kerbl20233d}, utilizing its core structure and functionalities. The project was built using Python and CUDA toolkit \cite{nickolls2008scalable}. Since 3D-GS requires a point cloud as input, we initialize $10^6$ points randomly to form the initial point cloud, as the structure-from-motion initialization \cite{schonberger2016structure} used in the original model is not suitable for event data. All experiments were performed on an NVIDIA RTX 3090 GPU.

We adhered to specific hyperparameter settings for performance and optimization. The total number of training iterations was set to 50,000. For position optimization, the initial and final learning rates were $1.6 \times 10^{-4}$ and $1.6 \times 10^{-6}$, respectively. The learning rates for feature, opacity, scaling, and rotation optimization were $2.5 \times 10^{-3}$, $5 \times 10^{-2}$, $5 \times 10^{-3}$, and $1 \times 10^{-3}$, respectively. Densification parameters included a percent density of $0.01$, a densification interval of $100$ iterations, and densification occurring between iterations $500$ and $50,000$, triggered by a gradient threshold of $2 \times 10^{-4}$. The opacity was reset every $3,000$ iterations. These settings were chosen from the default 3D-GS settings to balance training stability and performance.

\subsection{Real-World Macro Sequences}
\label{sec:real_world_experiment}

\begin{table}[ht]
\caption{Quantitative Comparison on Real-World Sequences. The top two results are marked with \textbf{bold} (1st) and \underline{underline} (2nd).}
\centering
\scalebox{.90}{
\begin{tabular}{c|ccccc}
\hline\hline
Metric & PSNR↑ & SSIM↑ & Training Time↓ & FPS↑ & Memory↓ \\ \hline\hline
Scene & \multicolumn{5}{c}{Toy House} \\ \hline
3D-GS & 23.8 & 0.80 & \textbf{7min} & \textbf{80.2} & \textbf{3GB} \\
EventNeRF & 24.8 & \underline{0.88} & 14h & 0.30 & 15GB \\
Ev-GS & 19.2 & 0.68 & 10min & 59.7 & \underline{5GB} \\
Event-3DGS & \underline{26.7} & 0.85 & 15min & 55.2 & 7GB \\
Ours & \textbf{28.7} & \textbf{0.90} & \underline{9min} & \underline{62.7} & \underline{5GB} \\ 
\hline\hline
Scene & \multicolumn{5}{c}{Optical Component} \\ \hline
3D-GS & 20.1 & \underline{0.83} & \textbf{7min} & \textbf{80.9} & \textbf{3GB} \\ 
EventNeRF & 24.1 & 0.81 & 14h & 0.33 & 15GB \\
Ev-GS & 20.2 & 0.70 & 10min & 55.3 & \underline{5GB} \\
Event-3DGS & \underline{25.8} & 0.82 & 15min & 53.0 & 7GB \\
Ours & \textbf{26.3} & \textbf{0.87} & \underline{9min} & \underline{63.0} & \underline{5GB} \\
\hline\hline
Scene & \multicolumn{5}{c}{Keyboard} \\ \hline
3D-GS & 25.3 & 0.86 & \textbf{7min} & \textbf{81.3} & \textbf{3GB} \\ 
EventNeRF & 25.0 & \underline{0.84} & 14h & 0.32 & 15GB \\
Ev-GS & 13.6 & 0.60& 10min & 61.2 & \underline{5GB} \\
Event-3DGS & \underline{25.2} & 0.79 & 15min & 57.0 & 7GB \\
Ours & \textbf{30.5} & \textbf{0.92} & \underline{9min} & \underline{62.3} & \underline{5GB} \\
\hline\hline
Scene & \multicolumn{5}{c}{Cloth} \\ \hline
3D-GS & 24.1 & 0.81 & \textbf{7min} & \textbf{81.8} & \textbf{3GB} \\ 
EventNeRF & 25.5 & \underline{0.86} & 14h & 0.32 & 15GB \\
Ev-GS & 11.6 & 0.58 & 10min & 55.9 & \underline{5GB} \\
Event-3DGS & \underline{26.4} & 0.77 & 15min & 53.2 & 7GB \\
Ours & \textbf{31.1} & \textbf{0.94} & \underline{9min} & \underline{61.3} & \underline{5GB} \\
\hline\hline
\end{tabular}}
\label{compare real}
\end{table}


In this experiment, we demonstrate the efficacy of our rendering method compared to others in real-world macro sequences. Due to the fast sweep during the imaging and data collection process, the captured frames are blurred. Training with blurred and low-quality frames, Traditional 3D-GS struggles, leading to blurred and indistinct renderings, as shown in part (b) in Figure \ref{fig:vis all}. Existing event-based approaches such as EventNeRF \cite{rudnev2023eventnerf}, Ev-GS \cite{wu2024EV-GS}, and Event-3DGS \cite{han2024event}, on the other hand, are mainly designed for synthetic event data. Therefore, when dealing with real-world event streams that are noisy and with more randomness, they failed to produce high-quality radiance fields and rendering results, shown in parts (c, d, and e) in Figure \ref{fig:vis all}. 
Our method, however, utilizes the captured event streams and the first static frame at the initial position of the camera. Event streams are, in nature, less sensitive to lighting changes. Therefore, with our designed approach, we can produce sharp and clear rendering results, as shown in sub-figure (f) in Figure \ref{fig:vis all}. Our method shows a sharp rendering result with clear details and semantics. The quantitative comparison of rendering results is shown in Table \ref{compare real}. Our method addresses the limitations of traditional 3D-GS and challenges faced by existing event-based radiance field approaches. The results demonstrate that our approach not only preserves but enhances the clarity and detail of the rendered images, providing a robust solution for real-world applications.

\subsection{Real-World Microscopic Sequences}
\label{Sec: real-world micro}
In microscopy settings, as demonstrated in Figure \ref{fig:setup} part (b), even small movements of the scene or camera can cause significant displacements, resulting in blurry imaging when using frame-based cameras. This limitation often hampers traditional radiance field rendering methods from reconstructing sharp and meaningful textural information in microscopy settings. In contrast, our approach leverages event streams for radiance field rendering, making it particularly suitable for such challenging microscopy settings. The quantitative comparison results are shown in Table \ref{compare real micro}, and our approach renders high-quality reconstruction results with decent PSNR, SSIM scores, while keeping a balanced efficiency in training and rendering.
As shown in the right part of Figure \ref{fig:vis all}, we present visual comparisons between our approach, Frame-Based 3D-GS, and existing event-based radiance field approaches.

\begin{figure*}[htbp]
\begin{center}
  \includegraphics[width=1 \linewidth]{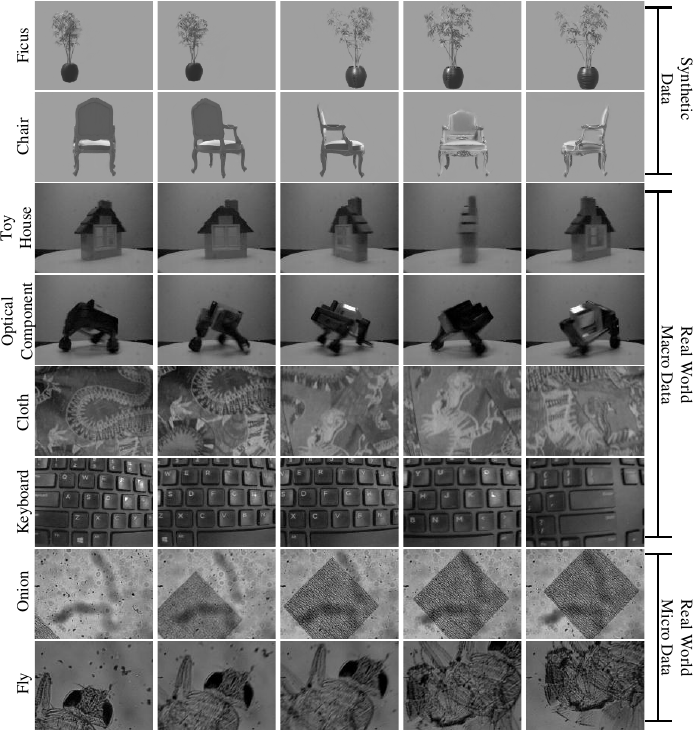}
\end{center}
   \caption{More visualization of the rendered results on objects from different imaging settings. The top two rows are synthetic data. The middle four rows are real-world macro objects. The bottom two rows are real-world objects captured under the microscope. We showcase the rendering results from different view perspectives for each object. We also demonstrate the rendered results on unseen views of the keyboard and cloth data. This also showcases that our approach is able to reconstruct sharp unseen views with clear semantic information.}
\label{fig:supp vis3}
\end{figure*}

We provide a visualization of different view angles from different samples across different imaging setups in Figure \ref{fig:supp vis3}. To demonstrate the capability of novel view synthesis, we randomly add a small displacement on the input coordinate (horizontal and vertical directions) to simulate unseen camera directions and render the unseen view based on these unseen coordinates.

\begin{table}[ht]
\caption{Quantitative Comparison on Real-World Microscopy Sequences. The top two results are marked with \textbf{bold} (1st) and \underline{underline} (2nd).}
\centering
\scalebox{.90}{
\begin{tabular}{c|ccccc}
\hline\hline
Metric & PSNR↑ & SSIM↑ & Training Time↓ & FPS↑ & Memory↓ \\ \hline\hline
Scene & \multicolumn{5}{c}{Onion} \\ \hline
3D-GS & 23.8 & 0.80 & \textbf{7min} & \textbf{80.2} & \textbf{3GB} \\
EventNeRF & \underline{24.8} & \underline{0.88} & 14h & 0.30 & 15GB \\
Ev-GS & 8.6 & 0.55 & 10min & 62.0 & \underline{5GB} \\
Event-3DGS & 18.9 & 0.75 & 15min & 57.0 & 7GB \\
Ours & \textbf{28.7} & \textbf{0.90} & \underline{9min} & \underline{62.7} & \underline{5GB} \\ 
\hline\hline
Scene & \multicolumn{5}{c}{Fly} \\ \hline
3D-GS & 20.1 & 0.83 & \textbf{7min} & \textbf{80.9} & \textbf{3GB} \\ 
EventNeRF & \underline{24.1} & 0.81 & 14h & 0.33 & 15GB \\
Ev-GS & 9.8 & 0.55 & 10min & 60.0 & \underline{5GB} \\
Event-3DGS & \underline{24.1} & \underline{0.82} & 15min & 55.0 & 7GB \\
Ours & \textbf{26.3} & \textbf{0.87} & \underline{9min} & \underline{63.0} & \underline{5GB} \\
\hline\hline
\end{tabular}}
\label{compare real micro}
\end{table}

\subsection{Synthetic Sequences}
\label{Sec: Synthetic}
\begin{figure}[t]
	
	\centering
	
	\includegraphics[width=\linewidth,scale=1.00]{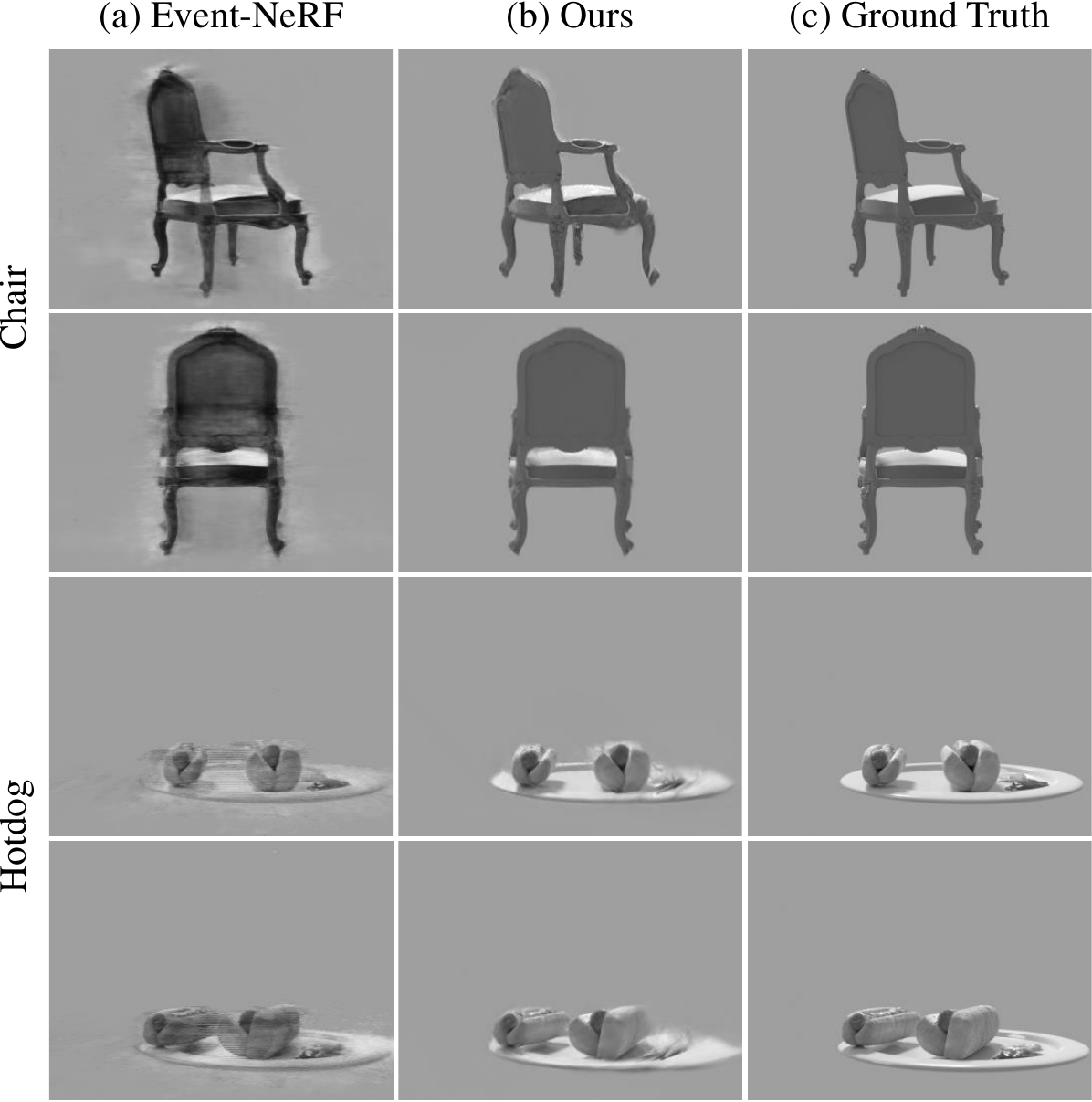}
	
	\caption{Visual Comparison between our approach and \cite{rudnev2023eventnerf}}
	
	\label{fig:vis synthesis}
	
\end{figure}
As previously mentioned, we evaluated synthetic sequences sourced from \cite{mildenhall2021nerf}, which include various scenes such as chairs, hot dogs, ficus, and mic. Table \ref{compare event nerf} presents a quantitative comparison with EventNeRF \cite{rudnev2023eventnerf} and Ev-GS \cite{wu2024EV-GS} \footnote{Comparison methods are evaluated with 360-degree data and the score was reported from the original paper, whereas our approach is evaluated on 90-degree data to replicate the sweep.}. Our method demonstrates significant advantages over EventNeRF across these scenes. Specifically, in terms of PSNR and SSIM, as shown in Table \ref{compare event nerf}, our approach consistently achieves higher values, indicating superior reconstruction fidelity and better structural similarity between the reconstructed and ground truth images. Notably, our method also boasts significantly reduced training times, requiring only approximately 10 minutes compared to EventNeRF's 14 hours for all scenes. This accelerated training time enables more efficient model development and experimentation. Moreover, our method achieves a higher temporal resolution of around 60 fps, compared to EventNeRF's 0.32 fps. This high frame rate is crucial for smooth visual experiences, reducing motion sickness and enhancing productivity and competitiveness. Lastly, our method consumes less memory during the rendering stage, utilizing only 5 GB compared to EventNeRF's 15 GB, which is advantageous for memory-constrained environments and facilitates scalability.

Qualitatively, we further compare visual results for two sequences. As shown in Figure \ref{fig:vis synthesis}, our Ev-GS effectively captures view-dependent effects, structures, and intensity in the frames compared to previous methods.

\begin{table}[h]
\caption{Quantitative Comparison Against Event-NeRF \cite{rudnev2023eventnerf} and Ev-GS \cite{wu2024EV-GS} on Synthetic Sequences.}
\centering
\scalebox{.9}{
\begin{tabular}{c|ccccc}
\hline\hline
Metric & PSNR↑ & SSIM↑ & Training Time↓ & FPS↑ & Memory↓ \\ \hline\hline
Scene & \multicolumn{5}{c}{Chair} \\ \hline
EventNeRF & 25.6 & 0.91 & 14h & 0.32 & 15GB \\
Ev-GS & 28.1 & 0.93 & 9min & 53.1 & \textbf{5GB} \\
Ours & \textbf{31.7} & \textbf{0.94} & \textbf{8min} & \textbf{65.7} & \textbf{5GB} \\ \hline\hline
Scene & \multicolumn{5}{c}{Ficus} \\ \hline
EventNeRF & 27.1 & 0.91 & 14h & 0.33 & 15GB \\
Ev-GS & 28.1 & 0.92 & 9min & 63.0 & \textbf{5GB} \\
Ours & \textbf{29.1} & \textbf{0.93} & \textbf{8min} & \textbf{65.9} & \textbf{5GB} \\ \hline\hline
Scene & \multicolumn{5}{c}{Hotdog} \\ \hline
EventNeRF & 26.0 & 0.92 & 14h & 0.32 & 15GB \\
Ev-GS & 25.7 & 0.93 & 12min & 55.3 & \textbf{5GB} \\ 
Ours & \textbf{29.3} & \textbf{0.94} & \textbf{8min} & \textbf{65.3} & \textbf{5GB} \\ \hline\hline
Scene & \multicolumn{5}{c}{Mic} \\ \hline
EventNeRF & 25.0 & 0.91 & 14h & 0.32 & 15GB \\
Ev-GS & 24.5 & 0.92 & 11min & \textbf{66.0} & \textbf{5GB} \\
Ours & \textbf{27.5} & \textbf{0.93} & \textbf{8min} & 64.0 & \textbf{5GB} \\ \hline\hline
\end{tabular}}
\label{compare event nerf}
\end{table}

\subsection{Ablation Study}
\label{Sec: Ablation}
\subsubsection{End-To-End Against Two Stages}
\begin{figure}[h]
	
	\centering
	
	\includegraphics[width=\linewidth,scale=1.00]{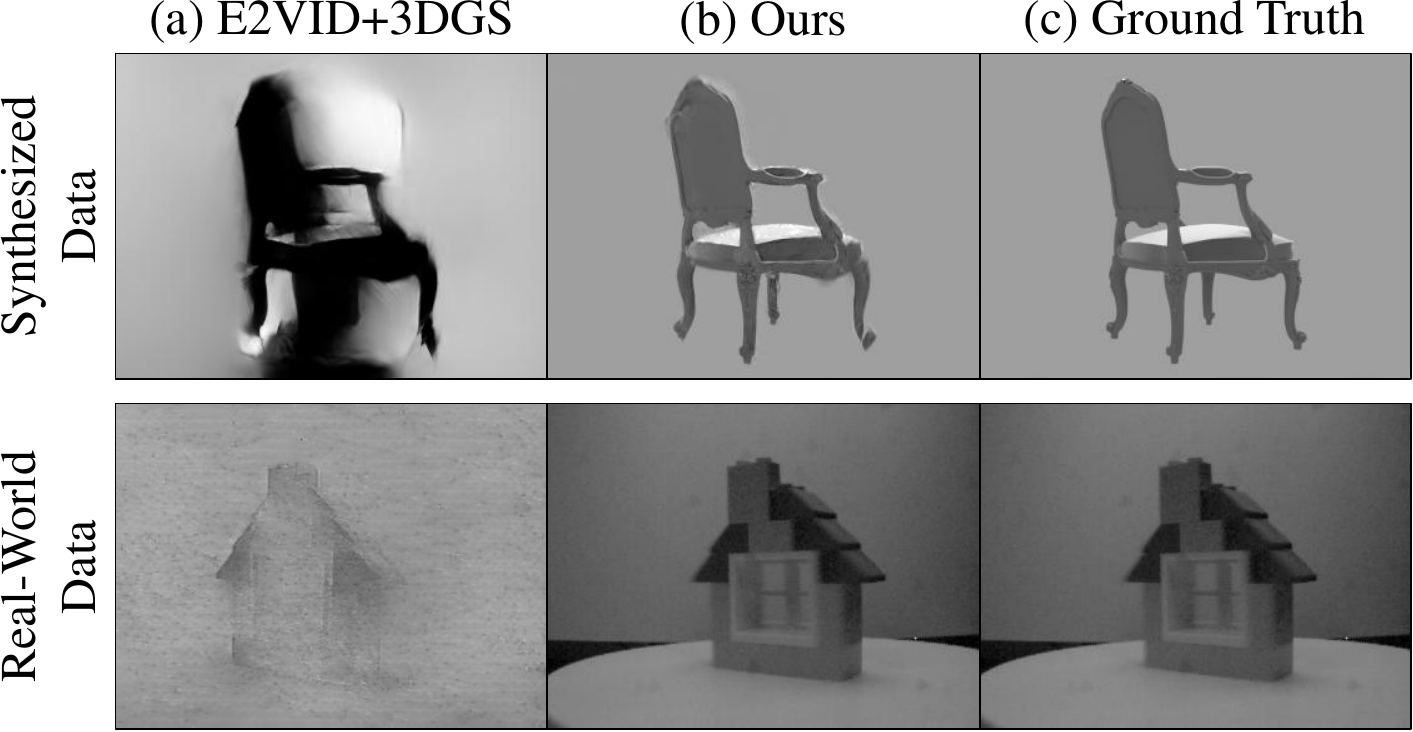}
	
	\caption{Visual Comparison between our end-to-end approach and Two-Stage baseline. (a) Use E2VID \cite{rebecq2019high} to generate training frames for 3D-GS; (b) the rendered results of our end-to-end approach; (c) the ground truth frame. We showcase this on both synthetic data and real-world data.}
	
	\label{fig:ablation}
	
\end{figure}

To assess the necessity and effectiveness of our proposed end-to-end design, we conducted an ablation study. Specifically, we compared our rendering results to a two-stage baseline approach: first converting event data to frame data using E2VID \cite{rebecq2019high} and then applying 3D-GS for rendering. Table \ref{table: Ablation} summarizes the quantitative evaluation results for two different scenes, Chair and Ficus, based on PSNR and SSIM metrics. As shown in Table \ref{table: Ablation}, our end-to-end design consistently outperforms the two-stage E2VID+3D-GS approach across both scenes and evaluation criteria. The limitations of the E2VID model become apparent when confronted with frames unseen during training. Since E2VID relies on the frames it was trained on, its intermediate results may become inaccurate when presented with novel frames. This inaccuracy propagates through the subsequent 3D-GS stage, resulting in poorer rendering quality, as illustrated in Figure \ref{fig:ablation}. These results underscore the necessity and effectiveness of our end-to-end design, seamlessly integrating event-based video input processing with 3D-GS reconstruction. This integration leads to significantly improved reconstruction quality and fidelity compared to the two-stage approach.

\begin{table}[h]
\caption{Ablation Study on our end-to-end approach against the two-stage baseline of using \cite{rebecq2019high} plus 3D-GS. Both evaluation metrics indicates that our end-to-end approach is more effective and necessary for training and rendering across both synthesized and real-world data.}
\centering
\scalebox{1}{
\begin{tabular}{c|cc|cc}
\hline \hline
Scene & \multicolumn{2}{c}{Synthesized
} & \multicolumn{2}{c}{Real-World
} \\ \hline
Criteria $|$ Metric & PSNR↑ & SSIM↑ & PSNR↑ & SSIM↑ \\ \hline
E2VID+3D-GS & 15.2 & 0.72 & 13.2 & 0.68 \\
Ours & \textbf{29.4} & \textbf{0.94} & \textbf{29.2} & \textbf{0.91} \\
\hline \hline
\end{tabular}}
\label{table: Ablation}
\end{table}

\subsubsection{How Event Noise Affects Radiance Field Reconstruction}
Designing an approach that is robust across different real-world imaging settings is not trivial. Different hardware imaging systems introduce various challenges due to their unique characteristics and limitations. For instance, real-world macro imaging settings might deal with varying lighting conditions, while microscopy imaging settings often face higher levels of noise \cite{yang2023sci}. These variations make it difficult to maintain consistent performance across different scenarios. To address these challenges, we introduced the y-noise filter \cite{feng2020event} to eliminate event noise. This simple yet effective modification enhanced the overall performance of our method.

\begin{figure}[h]
	\centering
	\includegraphics[width=\linewidth, scale=1.0]{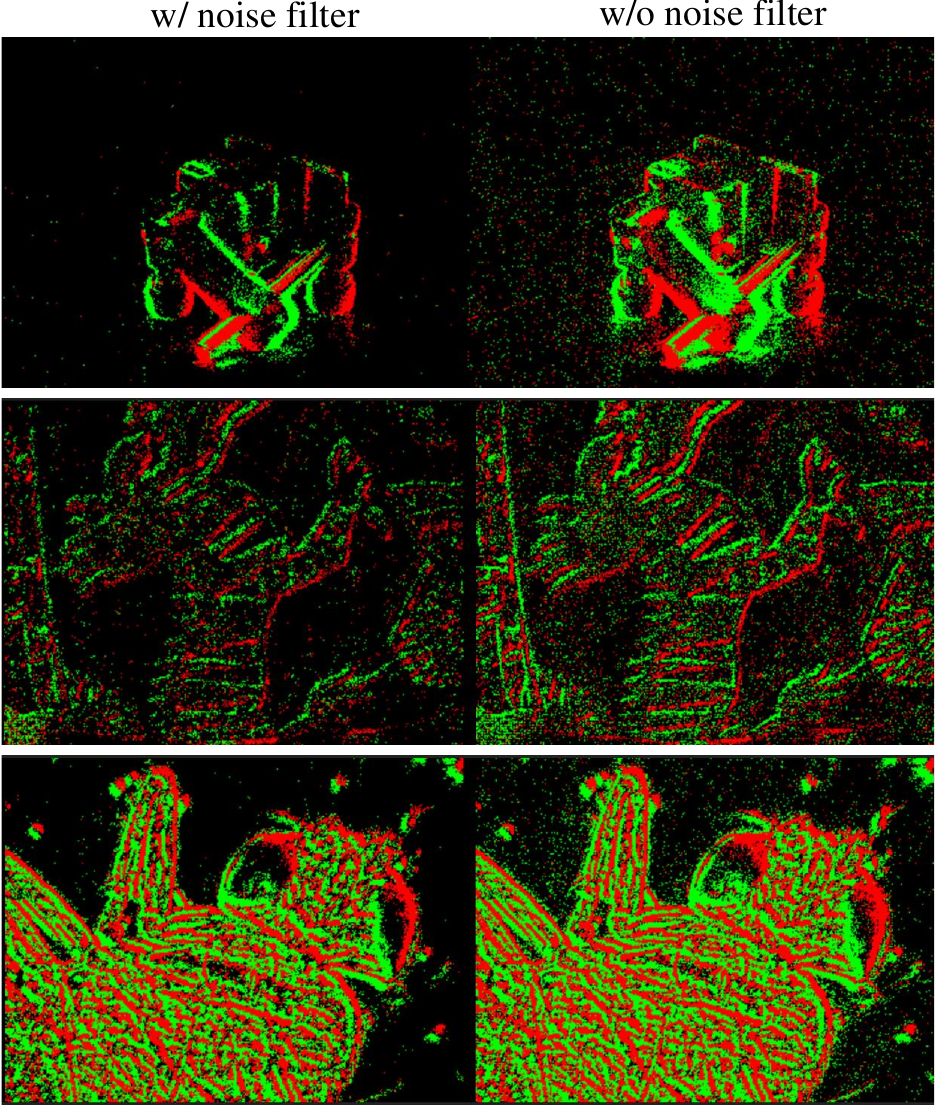}
	\caption{Visual comparison between noise-filtered events (left) and raw events captured (right). Raw events, especially triggered under different imaging settings, contain different noise and randomness patterns. To adapt Sweep-EvGS for data from different settings, noise is needed to be filtered and balanced.}
	\label{fig:ablation noise}
\end{figure}

In our experiments, we compared the performance of our method with and without the noise filter. Qualitatively, the input event is more clean on noncritical regions, indicated by Table \ref{fig:ablation noise}. An input event stream with less noise leads to better reconstruction of radiance field representations. The quantitative results are summarized in Table \ref{compare_mean}. The metrics used for evaluation are the Peak Signal-to-Noise Ratio (PSNR) and Structural Similarity Index (SSIM), both of which are standard measures for assessing image quality. The results indicate that incorporating a noise filter significantly improves the rendering performance. For synthetic data, the PSNR increased from 27.8 to 29.4, and the SSIM improved from 0.90 to 0.94. Similarly, the PSNR increased from 26.9 to 29.2 for real-world data, and the SSIM improved from 0.88 to 0.91. These improvements highlight the effectiveness of our approach in enhancing the quality of the rendered radiance fields, thereby making our method more robust across different types of data.

\begin{table}[h]
\caption{Impact of noise filtering on the rendering performance for synthetic and real-world data. Incorporating a noise filter significantly improves both PSNR and SSIM values.}
\centering
\setlength{\tabcolsep}{5mm}{
    \begin{tabular}{c|cc}
        \hline\hline
        Metric & PSNR↑ & SSIM↑ \\
        \hline\hline
        Data Type & \multicolumn{2}{c}{Synthetic} \\
        \hline
        w/o filter & 27.8 & 0.90 \\
        w noise filter & \textbf{29.4} & \textbf{0.94} \\
        \hline\hline
        Data Type & \multicolumn{2}{c}{Real-World} \\
        \hline
        w/o filter & 26.9 & 0.88 \\
        w noise filter & \textbf{29.2} & \textbf{0.91} \\
        \hline\hline
    \end{tabular}
}
\label{compare_mean}
\end{table}

\subsubsection{Robust Loss Function across Various Real-World Imaging Settings}
In this ablation study, we compare the performance of our proposed loss function against the traditional Mean Squared Error (MSE) loss. Our loss function incorporates a more sophisticated approach to supervising the rendering process. Incorporating normalization techniques into our loss function is essential for maintaining stability across various imaging conditions. Normalization helps to mitigate the impact of differing data distributions and scaling effects, ensuring that the loss function behaves consistently regardless of the specific imaging environment. This stability is particularly important for preserving the quality of rendering results, as it allows the model to perform reliably and effectively across diverse data sources and imaging settings. By stabilizing the loss, our approach ensures that high-quality results are achieved consistently, whether the data comes from synthetic or real-world imaging scenarios. In contrast, the traditional MSE loss focuses solely on minimizing the pixel-wise squared differences between predicted and ground truth images. This simple metric often fails to capture structural and perceptual nuances, which can result in lower visual quality.

Table \ref{ablation loss} summarizes the quantitative comparison between our loss function and the traditional MSE loss for both synthetic and real-world data. As illustrated, our loss function outperforms the MSE loss significantly. For synthetic data, the PSNR improves from 25.2 to 29.4 and the SSIM from 0.78 to 0.94. Similarly, for real-world data, the PSNR increases from 25.7 to 29.2 and the SSIM from 0.81 to 0.91. These results highlight the effectiveness of our loss function in capturing and preserving important visual and structural details, leading to higher rendering quality compared to the traditional MSE loss. The improved performance underscores the advantages of our approach in producing more accurate and visually appealing results.

\begin{table}[h]
\caption{Quantitative comparison of rendering performance using traditional MSE loss versus our proposed loss function. Our loss function significantly improves PSNR and SSIM metrics for both synthetic and real-world data.}
\centering
\setlength{\tabcolsep}{5mm}{
    \begin{tabular}{c|cc}
        \hline\hline
        Metric & PSNR↑ & SSIM↑ \\
        \hline\hline
        Data Type & \multicolumn{2}{c}{Synthetic} \\
        \hline
        MSE Loss & 25.2 & 0.78 \\
        Our Loss & \textbf{29.4} & \textbf{0.94} \\
        \hline\hline
        Data Type & \multicolumn{2}{c}{Real-World} \\
        \hline
        MSE Loss & 25.7 & 0.81 \\
        Our Loss & \textbf{29.2} & \textbf{0.91} \\
        \hline\hline
    \end{tabular}
}
\label{ablation loss}
\end{table}

\subsection{Qualitative Analysis}
\label{Sec: analysis}
As shown above, we provide rendering results of different kinds of sequences, demonstrating the different capabilities of our approach. Here, we provide more detailed analyses of the unique capability of our approach.

\begin{itemize}
    \item \textbf{Reconstruction of Sharpe Frames}: The first sequence involves a toy house. Traditional frame-based cameras capture blurred frames under fast camera movements, leading the traditional 3D-GS method to learn and render significant blurring, making it difficult to discern the object's details, as shown in the Toy House sequence in Figure \ref{fig:vis all}. In contrast, our method effectively uses the event stream data to reconstruct a sharp and well-defined image of the toy house. This highlights our method's capability to maintain clarity and detail, even in challenging imaging environments.
    
    \item \textbf{Reconstruction of Challenging Surface Materials}: The second sequence, shown in the introduction of the main text, showcases an optical component with vitric material that is reflective of lights. Traditional methods often struggle to capture the true geometry and appearance of reflective objects accurately. However, our approach excels in rendering these difficult materials, producing clear and accurate reconstructions.

    \item \textbf{Reconstruction of Semantic Information}: The third sequence, shown in the Keyboard sequence in Figure \ref{fig:vis all}, is the classical keyboard scene. We demonstrate this challenging sequence to showcase that our method is not only capable of reconstructing texture information but, compared to existing methods, is also more capable of reconstructing semantic information with clear and detailed visual quality.

    \item \textbf{Reconstruction of Complex Texture:} The fourth sequence, shown in Figure \ref{fig:supp vis3}, presents a cloth with complex textures. Our approach provides a high-fidelity reconstruction that is particularly challenging for event-based radius field rendering. Unlike conventional methods that may smooth over fine details, our method retains the rich detail and variability inherent in such textures. This capability is crucial for rendering applications in any field requiring accurate representation of complex patterns and materials.
\end{itemize}

\section{Conclusion and Discussion}
This paper introduces SweepEvGS, a novel method for robust and accurate novel view synthesis using an event camera. By harnessing the unique advantages of event cameras, SweepEvGS surpasses the limitations of traditional methods that rely on dense and high-quality training frames. It enables efficient novel view synthesis across various imaging settings with minimal data collection requirements. Experimental results demonstrate that SweepEvGS outperforms existing methods in terms of visual rendering quality, rendering speed, and computational efficiency across synthetic data, real-world scenarios, and microscopy settings. By leveraging event streams, SweepEvGS achieves sharper and clearer reconstructions, highlighting its effectiveness in challenging imaging environments. Overall, SweepEvGS represents a significant advancement in novel view synthesis techniques, offering practical solutions for dynamic and varied imaging applications.

\section{Limitations and Future Work}
Although SweepEvGS has shown effectiveness in novel view synthesis across different imaging setups with high efficiency, the data collection process is less adaptable. This limitation arises from the reliance on precise camera pose and extrinsic coordinate information for 3D-GS methods. Currently, our real data collection can be performed on machines that can transform the event camera with a known trajectory and speed, making it possible and easy to simulate coordinates. However, an ideal industrial-level algorithm should enable sweeps to be performed by human hands, with flexibility in trajectory and speed, while still producing high-quality view synthesis and reconstruction. A straightforward approach might be to utilize event-based stereo vision to generate coordinates, thereby making the process independent of specific hardware configurations. Future work could focus on further refining this technology to achieve the goal of a flexible, high-quality view synthesis and reconstruction system that is not tied to a particular hardware setup.

\section{Acknowledgments}
This work is supported by the Theme-based Research Scheme (Grant No. T45-701/22-R), National Natural Science Foundation of China (Grant No. 62136001, 62088102), Beijing Natural Science Foundation (Grant No. L233024), and Beijing Municipal Science \& Technology Commission, Administrative Commission of Zhongguancun Science Park (Grant No. Z241100003524012). 




 
%
\bibliographystyle{IEEEtran}
\bibliography{Ref}

\newpage

\section{Biography Section}
\vskip -25pt plus -1fil
\begin{IEEEbiography}[{\includegraphics[width=1in,height=1.25in,clip,keepaspectratio]{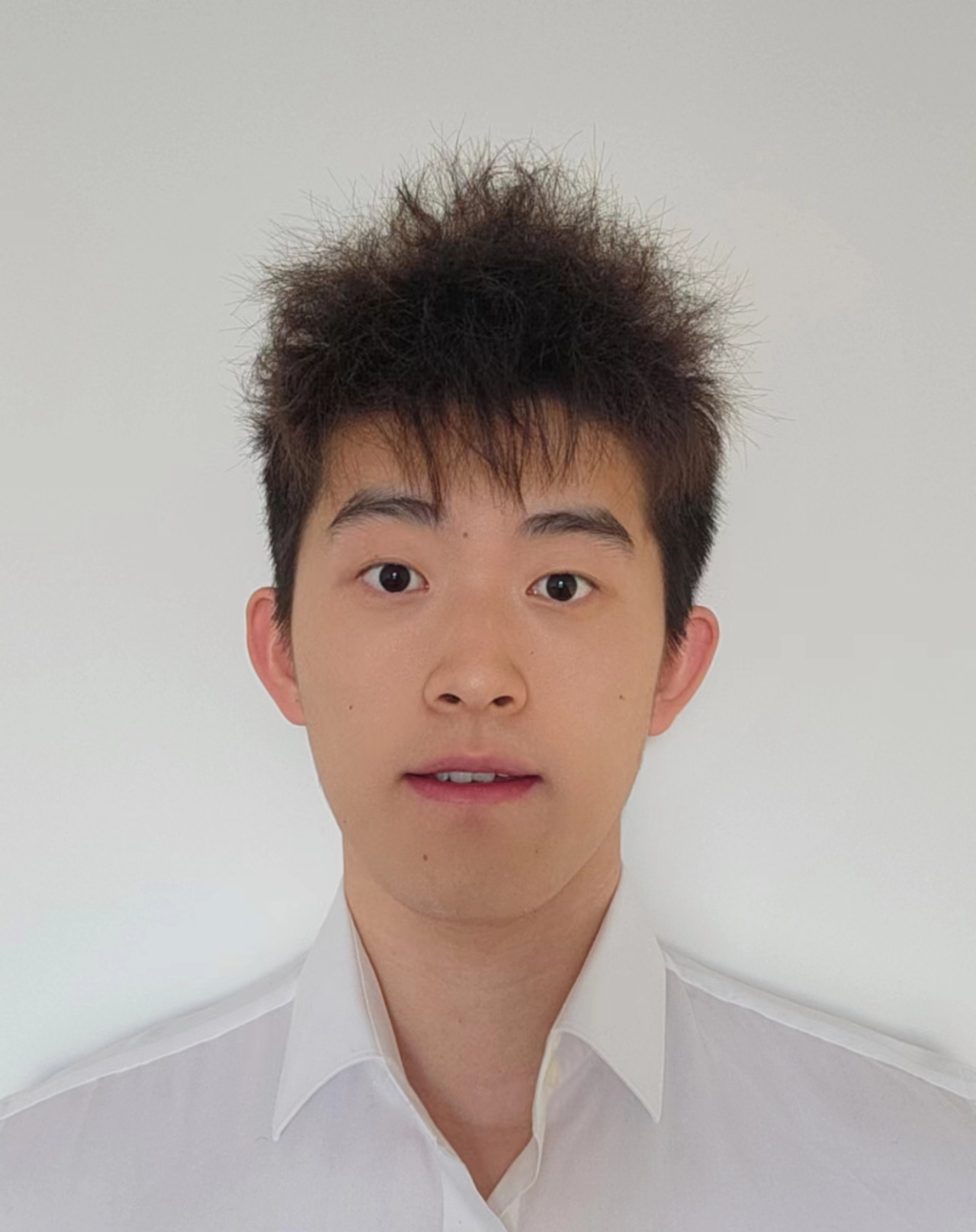}}]{Jingqian Wu} received the B.S. degree from Wake Forest University, USA, in 2022, and the M.S. degree from Northwestern University, USA, in 2024. He is currently pursuing the Ph.D. degree with the Department of Electrical and Electronic Engineering, The University of Hong Kong. His research interests include computational imaging, computer vision and event-based vision.
\end{IEEEbiography}
\vskip -25pt plus -1fil
\begin{IEEEbiography}[{\includegraphics[width=1in,height=1.25in,clip,keepaspectratio]{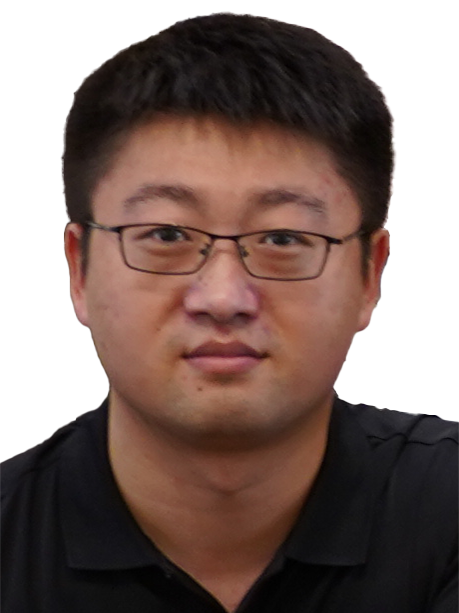}}]{SHUO ZHU} received his B.S. degree from the Changchun University of Science and Technology in 2016, M.S. degree from the University of Shanghai for Science and Technology in 2019, and Ph.D. degree in optical engineering from the Nanjing University of Science and Technology in 2023. He is now a postdoctoral fellow at the University of Hong Kong. His research interest is computational neuromorphic imaging and its optical applications.
\end{IEEEbiography}
\vskip -25pt plus -1fil
\begin{IEEEbiography}[{\includegraphics[width=1in,height=1.25in,clip,keepaspectratio]{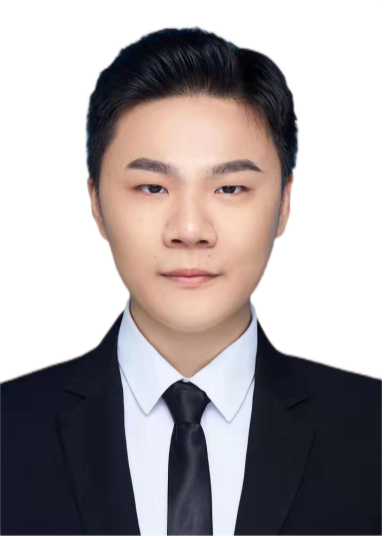}}]{CHUTIAN WANG} received the B.S. degree in Huang Kun Elite Class from the University of Science \& Technology Beijing in 2020, and the M.S. degree in the major of Optics and Photonics in Imperial College London in 2021. He was a research assistant in Zhejiang University until 2022. He is currently working towards his PhD degree with the Department of Electrical and Electronic Engineering, University of Hong Kong. His research interests include computational imaging and neuromorphic imaging.
\end{IEEEbiography}
\vskip -25pt plus -1fil
\begin{IEEEbiography}[{\includegraphics[width=1in,height=1.25in,clip,keepaspectratio]{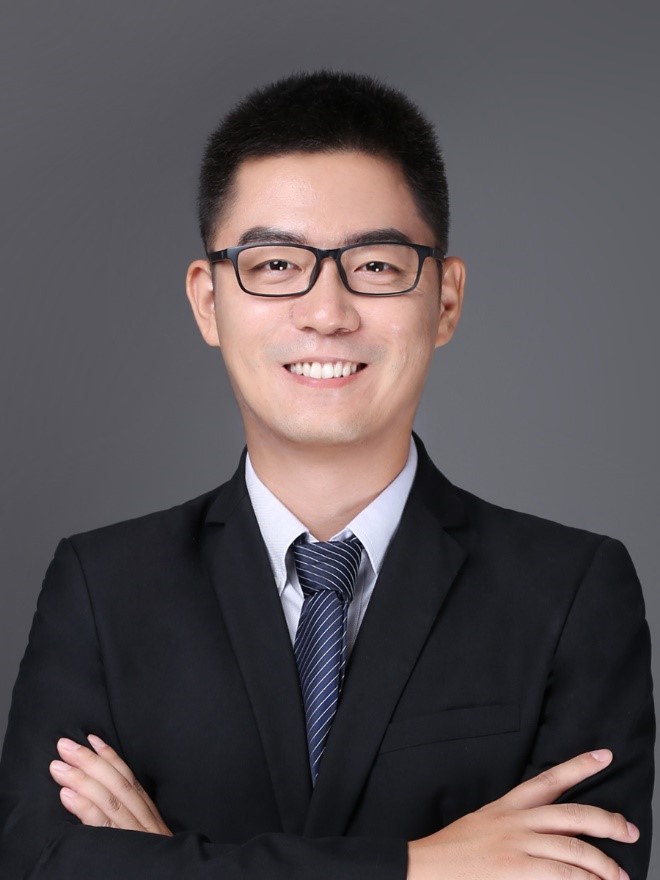}}]{Boxin Shi} (Senior Member, IEEE) received the B.E. degree from the Beijing University of Posts and Telecommunications, the M.E. degree from Peking University, and the PhD degree from the University of Tokyo, in 2007, 2010, and 2013. He is currently a Boya Young Fellow Associate Professor (with tenure) and Research Professor at Peking University, where he leads the Camera Intelligence Lab. Before joining PKU, he did research with MIT Media Lab, Singapore University of Technology and Design, Nanyang Technological University, National Institute of Advanced Industrial Science and Technology, from 2013 to 2017. His papers were awarded as Best Paper, Runners-Up at CVPR 2024, ICCP 2015 and selected as Best Paper candidate at ICCV 2015. He is an associate editor of TPAMI/IJCV and an area chair of CVPR/ICCV/ECCV. He is a senior member of IEEE.
\end{IEEEbiography}
\vskip -25pt plus -1fil
\begin{IEEEbiography}[{\includegraphics[width=1in,height=1.25in,clip,keepaspectratio]{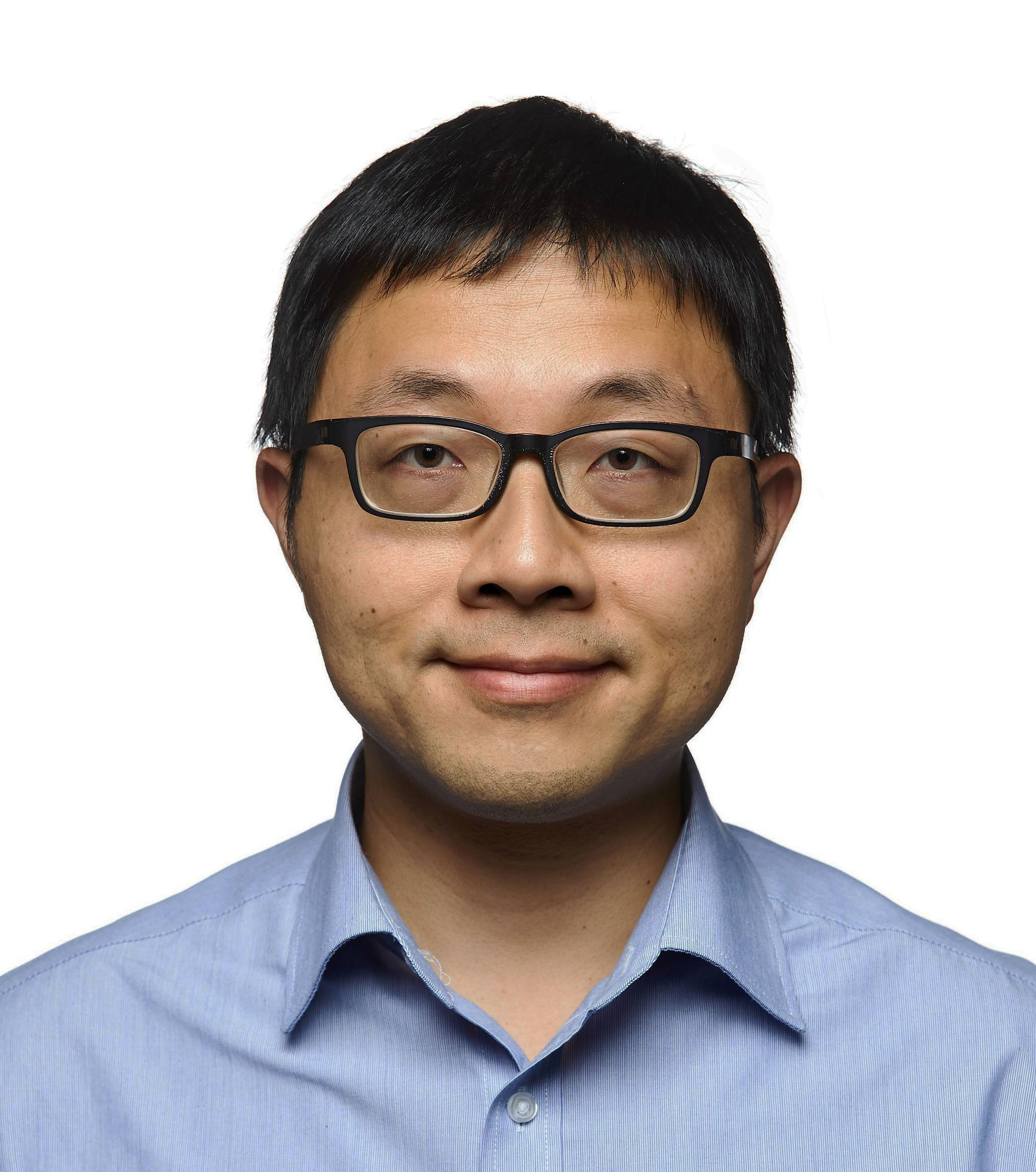}}]{EDMUND Y. LAM} (Fellow, IEEE) received the B.S., M.S., and Ph.D. degrees in electrical engineering from Stanford University. He was a Visiting Associate Professor with the Department of Electrical Engineering and Computer Science, Massachusetts Institute of Technology. He is currently a Professor of electrical and electronic engineering at The University of Hong Kong. He also serves as the Computer Engineering Program Director and a Research Program Coordinator with the AI Chip Center for Emerging Smart Systems. His research interest includes computational imaging algorithms, systems, and applications. He is a fellow of Optica, SPIE, IS\&T, and HKIE, and a Founding Member of the Hong Kong Young Academy of Sciences.
\end{IEEEbiography}

\vfill

\end{document}